\documentclass{article}

\PassOptionsToPackage{numbers, compress}{natbib}
\usepackage[final]{nips_2018}

\usepackage[utf8]{inputenc} 
\usepackage[T1]{fontenc}    
\usepackage{hyperref}       
\usepackage{url}            
\usepackage{booktabs}       
\usepackage{amsfonts}       
\usepackage{nicefrac}       
\usepackage{microtype}      
\usepackage{todonotes}
\usepackage{amsmath, amssymb}
\usepackage{subcaption}
\usepackage{enumitem}
\newcommand{\given}{\,|\,}

\newcommand{\bx}{\boldsymbol{x}}

\title{Measuring the Stability of EHR- and \\ EKG-based Predictive Models}

\author{
  Andrew C.~Miller\thanks{\texttt{am5171@columbia.edu}, \url{http://andymiller.github.io/}} \\
  Data Science Institute \\
  Columbia University \\
  \And
  Ziad Obermeyer \\
  School of Public Health \\
  University of California at Berkeley \\
  \And 
  Sendhil Mullainathan \\
  Booth School of Business \\
  University of Chicago \\
}

\begin{document}

\maketitle

\begin{abstract}
Databases of electronic health records (EHRs) are increasingly used to inform clinical decisions.
Machine learning methods can find patterns in EHRs that are predictive of future adverse outcomes.
However, statistical models may be built upon patterns of health-seeking behavior that vary across patient subpopulations, leading to poor predictive performance when training on one patient population and predicting on another. 
This note proposes two tests to better measure and understand model generalization.
We use these tests to compare models derived from two data sources: (i) historical medical records, and (ii) electrocardiogram (EKG) waveforms.
In a predictive task, we show that EKG-based models can be more stable than EHR-based models across different patient populations. 
\end{abstract}

\section{Introduction}
Patient history-based risk scores can inform physician decision-making and alter the course of treatment. 
The Framingham risk score for long-term coronary heart disease is one such an example \cite{wilson1998prediction}. 
In the past decade, large volumes of observational health care data --- administrative claims, digitized physician notes, and laboratory test values among others --- are now commonly used to build predictive models for a range of clinical purposes, including tracking disease progression \citep{cruz2006applications}, predicting near-term ICU interventions \citep{ghassemi2017predicting} or the onset of sepsis \citep{henry2015targeted}, and are incorporated into standardized prediction frameworks \citep{reps2018design}. 

Patterns in observational health data, by construction, reflect health-seeking behavior in the population served.
Consequently, predictive accuracy of statistical models can vary significantly if the underlying patterns of patient behavior vary in smaller groups within the population \citep{chen2018my}.
Further, under-served populations will, by definition, have less historical data on which to base predictive algorithms. 
Waveform and image data, such as electrocardiograms (EKGs) and echocardiograms, are a complementary source of information that can be used to build models of patient risk. 
EKGs, for instance, measure a patient's cardiac function, which may be predictive of future disease.
Patterns in a patient's EKG (which directly measures cardiac activity) that are predictive of heart failure may be more \emph{portable} than patterns in patient medical records across patient populations. 

In this note, we study the generalization performance of predictive algorithms trained on one population and tested on another.
We propose two statistics for measuring feature portability. 
As a case study, we construct models that predict the outcome of common lab test used to confirm heart attack --- troponin levels --- using (i) past diagnoses and medications and (ii) electrocardiogram waveform data. 
We show that EKG-based predictors are more stable across two sub-populations of patients --- those that use the health care system with high frequency and those that use the system with lower frequency. 
For each prediction algorithm (e.g.~data-source and training population), we examine the sources of distribution shift that can lead to poor generalization. 
We conclude with a discussion of future research directions. 

\section{Prediction and Conditional Stability}
Our goal is to construct an accurate predictive model for some outcome $y \in \{0, 1\}$ (e.g.~an adverse cardiac event) given some set of patient-specific data $\bx \in \mathbb{R}^D$ (e.g.~historical medical records or raw EKG waveforms).
We observe a set of these data drawn from a population distribution $P$
\begin{align}
    \bx_n, y_n &\sim P  && \text{ population distribution }\\
    \mathcal{D}_{P,N} &\triangleq \{ \bx_n, y_n \}_{n=1}^N && \text{ $P$-distributed dataset of size $N$.}
\end{align}
We use these data to train a model that predicts the conditional probability of $y=1$ given $\bx$
\begin{align}
    m_{\mathcal{D}_{P,N}} &\leftarrow \text{train}(\mathcal{D}_{P,N})
\end{align}
where the model $m_{\mathcal{D}_{P,N}}(\bx) \triangleq P(y=1 | \bx)$ approximates the conditional distribution.
For example, if $\bx$ are medical record features, $m_{\mathcal{D}_{P,N}}$ may be a logistic regression model fit with maximum likelihood.
If $\bx$ are EKG waveforms, $m_{\mathcal{D}_{P,N}}$ may be a convolutional neural network.

\vspace{-.5em}
\paragraph{Model Portability} 
How well does $m_{\mathcal{D}_{P,N}}$ perform on another distribution, $\bx, y \sim Q$?
For example, the $Q$ may be another hospital, health care system, or patient sub-population.
For a specific model $m_{\mathcal{D}_P}$ (e.g.~logistic regression with EHR data) and a \emph{different} sub-population $Q$, we can measure cross-generalization with the area under the ROC curve
\begin{align}
    G^{(m)}(\mathcal{D}_P, \mathcal{D}_Q) &= \text{AUC}(m_{\mathcal{D}_P}, \mathcal{D}_{Q})
\end{align}
where an entry $\mathcal{D}_P, \mathcal{D}_Q$ denotes a model trained on population $P$ and tested on population $Q$.
The diagonal entries $G^{(m)}(\mathcal{D}_{P}, \mathcal{D}'_{P})$ measure the standard notion of generalization with no distribution shift --- predictive performance on held out data from the same distribution. 
The off-diagonal entries $G^{(m)}(\mathcal{D}_{P}, \mathcal{D}_{Q})$ measure performance under a new test distribution, a problem addressed by domain adaption techniques \cite{ben2010theory}.
When generalization performance for population $P$ and population $Q$ are different, a natural question is what is driving that gap?  What features are portable?

\vspace{-.5em}
\paragraph{Distribution shift} 
Why do predictive features $\bx$ fail to generalize? 
Consider a predictive model trained on data from population $P$, which we denote with shorthand $m_{P}(\bx)$.
The model $m_P(\bx)$ is just a function of covariates $\bx$.  
When $\bx$ is drawn from $P$, $m_{P}(\bx)$ and $y$ have a joint distribution, induced by $P(\bx, y)$. 
When $\bx$ is drawn from a different population $Q$, $m_{P}(\bx)$ and $y$ have a \emph{different} joint distribution, induced by $Q(\bx, y)$.
For shorthand, define $\hat y_P = m_{P}(\bx)$. 
The conditional distributions induced by $P$, $Q$, and $m_{P}(\bx)$ can tell us what is ``stable'' and ``unstable''
\vspace{-.5em}
\begin{itemize} \itemsep 0pt
\item $P( \hat y_P \given y)$ vs.~$Q(\hat y_P \given y)$ : covariate stability 
\item $P( y \given \hat y_P)$ vs.~$Q(y \given \hat y_P)$ : predictive stability
\end{itemize}
\vspace{-.5em}
If covariates are conditionally unstable, this suggests different patterns of health-seeking behavior (or physiological differences) can be driving the generalization gap. 
If outcomes $y$ are conditionally unstable, this suggests that similar patterns in $\bx$ do not suggest $y$ the same way in population $P$ and population $Q$ --- there is something special about the prediction function tuned to population $P$ that does not apply to population $Q$.

We summarize \emph{covariate stability} by estimating the distance between conditional distributions
\begin{align}
    CS(m_{P}, y, P, Q) &= \text{KS}\left( P(\hat y | y); Q(\hat y | y) \right) && \text{covariate stability}
    \label{eq:covariate-stability}
\end{align}  
where $\text{KS}(\cdot; \cdot)$ is the Kolmogorov-Smirnov distance between the two distributions --- we use a two-sample estimator on held out data for this distance \citep{stephens1974edf}.  Intuitively, this will be low when covariates from the same class have the same distribution. 

We summarize \emph{predictive stability} with the difference in conditional expectation of $y$ given $\hat y_P$
\begin{align}
PS(\hat y; m_{P}, y, P, Q)
    &= \mathbb{E}_{y, \hat y \sim P}\left[ y | \hat y \right] - \mathbb{E}_{y', \hat y' \sim Q}\left[ y' | \hat y' \right ]
    \label{eq:predictive-stability}
\end{align}
This defines a function of $\hat y$ --- for each value of $\hat y$ the conditional distribution of outcome $y$ may be similar or different between $Q$ and $P$.
We visualize an estimate of this whole function, and compute a one dimensional summary (by averaging over an equal mixture of the $P$ and $Q$ distributions).\footnote{The notation $\hat y_P$, $P(\hat y_P, y)$ and $Q(\hat y_P, y)$ can appear confusing --- there are really three distributions to keep in mind: the distribution of the model training set ($P$), and the two distributions being compared (in our examples $P$ again and $Q$.)}

\section{Experiments}
We build models that predict the outcome of a troponin lab test in a cohort of emergency department patients. 
A troponin test measures the troponin T or I protein levels, which are released when heart muscle is damaged during a heart attack.
Troponin lab tests are used to determine if a patient is currently or about to suffer a heart attack.
Based on the guidelines in \cite{newby2012accf}, we define high to be a value of .04 ng/mL or greater.
We label a patient ``positive'' ($y=1$) if they have received a troponin result greater than or equal to .04 ng/mL within the seven days following their emergency department admission.
A patient is labeled ``negative'' ($y=0$) if they received a troponin result less than .04 ng/mL within the same time period. 
An algorithm that reliably predicts troponin outcome in the near term can be useful for assessing a patient's risk of heart attack.

We study the stability of predictive models derived from two distinct sources of covariate ($\bx$) data: 
\vspace{-.5em}
\begin{itemize}[leftmargin=*] \itemsep 1pt
    \item \texttt{ehr}: all diagnoses and medications administered over the past year, derived from ICD-9 codes collected by a network of hospitals.  We represent a patient's one-year history with a sparse binary vector of 2{,}372 diagnoses and medications --- a 1 indicates that the diagnosis/medication was present in the past year.  We fit predictive models using \texttt{ehr} data with XGBoost \citep{Chen:2016:XST:2939672.2939785}, a gradient boosted decision tree algorithm, tuning tree depth using a validation set. 
    \item \texttt{ekg}: raw electrocardiogram waveforms from an EKG administered at the beginning of the emergency department visit.  For the \texttt{ekg} data, we use the convolutional residual network described in \cite{rajpurkar2017cardiologist} on the three ``long leads'' (II, V1, V5). We use stochastic gradient descent, keeping the model with the best predictive performance on a validation set. 
\end{itemize}
\vspace{-.5em}
Following the notation above, we study \texttt{ehr}- and \texttt{ekg}-based predictors over two populations:
\vspace{-.5em}
\begin{itemize}[leftmargin=*] \itemsep 1pt
    \item $P \triangleq $ \emph{high-use patients}: patients with more than 8 visits to any department within the network of hospitals in the year before the emergency department encounter.
    \item $Q \triangleq $ \emph{low-use patients}: patients with 8 or fewer such visits in the same time frame.
\end{itemize} \vspace{-.5em}
Summary statistics of each population are presented in Table~\ref{tab:summary-table}.
We compare models derived from four combinations: trained on $\{P, Q\}$ using features  $\{\texttt{ehr}, \texttt{ekg} \}$.

\begin{table}[t!]
    \centering
    \begin{minipage}[c]{.68\textwidth}
        \scalebox{.78}{
        \begin{tabular}{llrrrrrr}
\toprule
     &      & \multicolumn{3}{l}{trop} & \multicolumn{2}{l}{age} & frac.~female \\
     &      &  frac.~pos &  \#~pos &  \#~obs & mean &  std &       mean \\
healthcare use & split &       &      &      &      &      &            \\
\midrule
low & train & 0.119 &  832 & 7000 & 58.4 & 18.4 &      0.459 \\
     & val & 0.116 &  116 & 1000 & 58.4 & 18.2 &      0.447 \\
     & test & 0.120 &  276 & 2298 & 58.2 & 18.4 &      0.438 \\
high & train & 0.177 & 1236 & 7000 & 63.9 & 15.3 &      0.583 \\
     & val & 0.217 &  217 & 1000 & 62.2 & 17.1 &      0.533 \\
     & test & 0.210 &  941 & 4491 & 63.5 & 15.5 &      0.555 \\
\bottomrule
\end{tabular}

        }
    \end{minipage}%
    \begin{minipage}[c]{.31\textwidth}
        \caption{Data summary.  We train prediction functions using equally sized populations of low and high usage patients (7{,}000 train, 1{,}000 validation).  We report test statistics on held-out patients. }
        \label{tab:summary-table}
    \end{minipage}
    \vspace{-.25em}
\end{table}

\begin{figure}[t!]
    \centering
    \begin{minipage}[c]{.27\textwidth}
        \caption{Predictive performance of each classifier by AUC. 
        The test population is labeled on the horizontal axis; the train population is labeled by color --- blue indcates trained on low-use ($Q$) patients, and orange indicates trained on high-use ($P$) patients.
        }
        \label{fig:auc-generalization}
    \end{minipage}~%
    \begin{minipage}[c]{0.72\textwidth}
        \begin{subfigure}[b]{.49\linewidth}
            \centering
            \includegraphics[width=\textwidth]{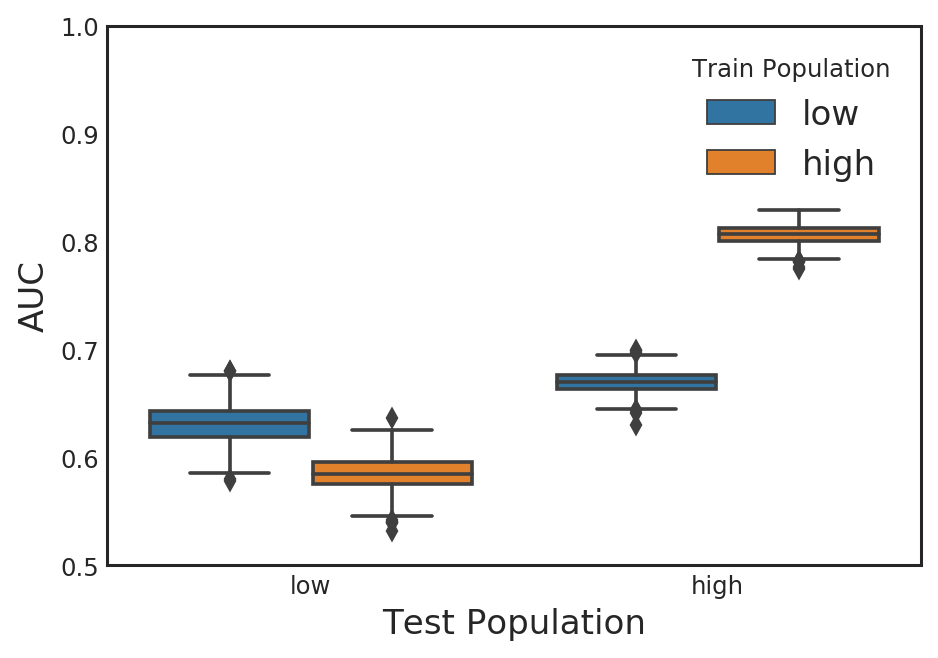}
            \caption{\texttt{ehr}-based models}
            \label{fig:ehr-auc}
        \end{subfigure}%
        \begin{subfigure}[b]{.49\linewidth}
            \centering
            \includegraphics[width=\textwidth]{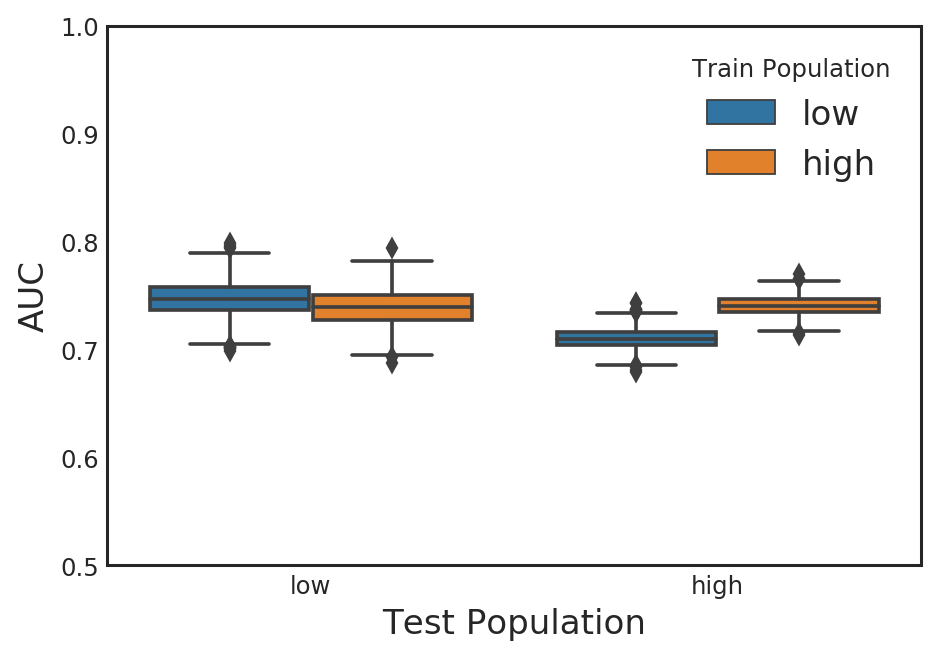}
            \caption{\texttt{ekg}-based models}
            \label{fig:ekg-auc}
        \end{subfigure}
    \end{minipage}
    \vspace{-.5em}
\end{figure}

\vspace{-.5em}
\paragraph{Generalization}
We present the \emph{within-distribution} generalization performance (e.g.~$G^{(m)}(P,P)$) and the \emph{between-distribution} performance (e.g.~$G^{(m)}(P, Q)$) in Figure~\ref{fig:auc-generalization}. 
The \texttt{ehr} model trained on the low-use ($Q$) population generalizes slightly better on new patients from $Q$ than the \texttt{ehr} model trained on the high-use ($P$) population.
The \texttt{ehr} model trained on $P$ (high-use) generalizes significantly better on patients from $P$ than the model trained on patients from $Q$, as summarized in Figure~\ref{fig:ehr-auc}.
On the other hand, we see that models trained on \texttt{ekg} data from either $P$ or $Q$ have similar predictive performance on both the $P$ and $Q$ distributions, as seen in Figure~\ref{fig:ekg-auc}.

\begin{figure}
    \centering
    \begin{minipage}[c]{.6\textwidth}
        \centering
        \includegraphics[width=\textwidth]{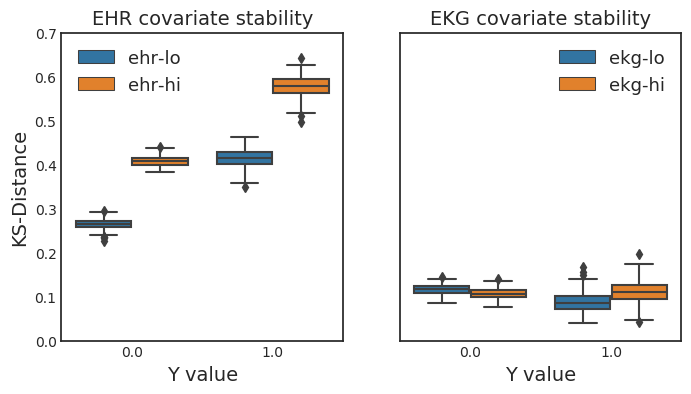}
    \end{minipage}~%
    \begin{minipage}[c]{.38\textwidth}
        \centering
        \caption{Distribution comparison summary of covariate stability.  
        The left plot compares models derived from EHR features. 
        Each entry estimates the statistical distance described in Equation~\ref{eq:covariate-stability} --- in words, for true value $y$ (and accompanying $\bx$, how different are the distributions $\hat y(\bx)$ when $\bx \sim P$ and $\bx \sim Q$?
        We see EKG features are far more stable under this measure (lower is better).}
        \label{fig:covariate-stability}
    \end{minipage}
    \vspace{-.5em}
\end{figure}

\vspace{-.5em}
\paragraph{Covariate stability}
Figure~\ref{fig:covariate-stability} shows the covariate stability statistics for the two sets of models, trained on the two populations. 
We can see that the covariate distributions $P(\hat y | y)$ and $Q(\hat y | y)$ are significantly closer when $\hat y$ is based on \texttt{ekg} features than \texttt{ehr} features. 
This strongly suggests that behavioral characteristics between high- and low-use patients \emph{with the same outcome} are driving a statistically significant difference in diagnosis and medication covariates $\bx$.
The \texttt{ekg} features, on the other hand, are significantly more stable between the high- and low-use patients.  
Intuitively, physiological measurements are not subject to the same biases that human behavior and decision-making introduce. 
The distributions of $\hat y$ given $y=0$ and $y=1$ for both the \texttt{ehr} and \texttt{ekg} models are depicted in Appendix Figure~\ref{fig:covariate-distributions}. 


\begin{figure}[t!]
    \centering
    \begin{minipage}[c]{.44\textwidth}
        \centering
        \caption{Predictive stability.  We summarize Equation~\ref{eq:predictive-stability} for \texttt{ehr}- (left) and \texttt{ekg}- (right) based models.  The vertical axis measures the difference between $E_Q[ y| \hat y] - E_P[y | \hat y]$ averaged over bins of $\hat y$.  Blue corresponds to models trained on low-use patients ($Q$) and orange to models trained on high-use patients ($P$).  For three models, we see a detectable bias, suggesting the functional relationship between $\bx$ and $y$ may differ slightly between groups. }
        \label{fig:predictive-stability-summary}
    \end{minipage}~%
    \begin{minipage}[c]{.54\textwidth}
        \centering
        \includegraphics[width=\textwidth]{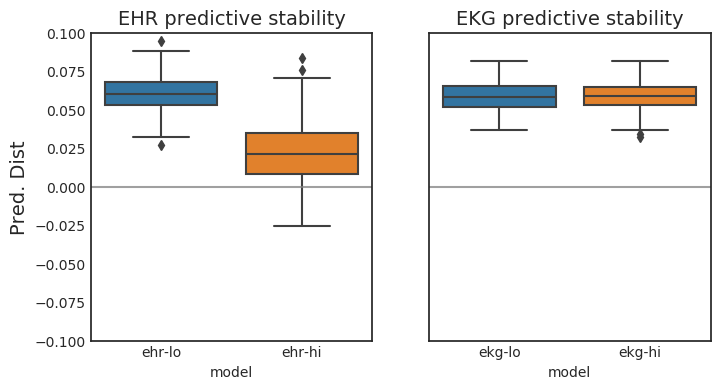}
    \end{minipage}
    \vspace{-.5em}
\end{figure}

\vspace{-.5em}
\paragraph{Predictive stability}
We present our one-dimensional predictive stability summary of Equation~\ref{eq:predictive-stability} in Figure~\ref{fig:predictive-stability-summary}, and depict full estimates in Figure~\label{fig:predictive-stability-full}. 
The average deviation between $E_Q[y | \hat y]$ and $E_P[y | \hat y]$ are similar for the \texttt{ekg} and \texttt{ehr} models and small, though generally non-zero (on the order of .05 to .08).
Both sets of models can make modest improvements, but this source of instability does not appear to be driving a large difference in generalization performance. 
More details are in Figure~\ref{fig:predictive-stability-full}. 

\section{Discussion}
We proposed two tests to better understand the portability of \texttt{ehr} and \texttt{ekg} features between patient populations.
Our analysis suggests that EKG features can be stable across low- and high-use patients, whereas diagnoses and medications are not. 
We see multiple avenues of future work. 
Given the observed covariate instability, we hope to model patient health-seeking behavior which lead to statistical differences in \texttt{ehr} data. 
We also hope that models for patterns of missing data can be effectively combined with \texttt{ekg} data to construct a predictor that is both accurate and portable. 

Additionally, it is presciently noted in \cite{chen2018my} that a strategy for making predictors more equitable is to pay the cost of additional (and focused) data collection.
Our analysis suggests a variant of this remedy --- find sources of data that are more likely to generalize across populations. 
Physician notes, for example, are likely to be influenced by physician-specific biases, but are still may be more portable than billing records --- a notion we hope to quantify and measure. 
EKGs and other physiological signals are not subject to the same patterns of presence and missingness in health record data, which is driven by unobserved human behavior and decision-making.

{\small
\bibliography{refs.bib}

\begin{thebibliography}{11}
\providecommand{\natexlab}[1]{#1}
\providecommand{\url}[1]{\texttt{#1}}
\expandafter\ifx\csname urlstyle\endcsname\relax
  \providecommand{\doi}[1]{doi: #1}\else
  \providecommand{\doi}{doi: \begingroup \urlstyle{rm}\Url}\fi

\bibitem[Wilson et~al.(1998)Wilson, D’Agostino, Levy, Belanger, Silbershatz,
  and Kannel]{wilson1998prediction}
Peter~WF Wilson, Ralph~B D’Agostino, Daniel Levy, Albert~M Belanger, Halit
  Silbershatz, and William~B Kannel.
\newblock Prediction of coronary heart disease using risk factor categories.
\newblock \emph{Circulation}, 97\penalty0 (18):\penalty0 1837--1847, 1998.

\bibitem[Cruz and Wishart(2006)]{cruz2006applications}
Joseph~A Cruz and David~S Wishart.
\newblock Applications of machine learning in cancer prediction and prognosis.
\newblock \emph{Cancer informatics}, 2:\penalty0 117693510600200030, 2006.

\bibitem[Ghassemi et~al.(2017)Ghassemi, Wu, Hughes, Szolovits, and
  Doshi-Velez]{ghassemi2017predicting}
Marzyeh Ghassemi, Mike Wu, Michael~C Hughes, Peter Szolovits, and Finale
  Doshi-Velez.
\newblock Predicting intervention onset in the icu with switching state space
  models.
\newblock \emph{AMIA Summits on Translational Science Proceedings},
  2017:\penalty0 82, 2017.

\bibitem[Henry et~al.(2015)Henry, Hager, Pronovost, and
  Saria]{henry2015targeted}
Katharine~E Henry, David~N Hager, Peter~J Pronovost, and Suchi Saria.
\newblock A targeted real-time early warning score (trewscore) for septic
  shock.
\newblock \emph{Science translational medicine}, 7\penalty0 (299):\penalty0
  299ra122--299ra122, 2015.

\bibitem[Reps et~al.()Reps, Schuemie, Suchard, Ryan, and
  Rijnbeek]{reps2018design}
Jenna~M Reps, Martijn~J Schuemie, Marc~A Suchard, Patrick~B Ryan, and Peter~R
  Rijnbeek.
\newblock Design and implementation of a standardized framework to generate and
  evaluate patient-level prediction models using observational healthcare data.
\newblock \emph{Journal of the American Medical Informatics Association}.

\bibitem[Chen et~al.(2018)Chen, Johansson, and Sontag]{chen2018my}
Irene Chen, Fredrik~D Johansson, and David Sontag.
\newblock Why is my classifier discriminatory?
\newblock \emph{arXiv preprint arXiv:1805.12002}, 2018.

\bibitem[Ben-David et~al.(2010)Ben-David, Blitzer, Crammer, Kulesza, Pereira,
  and Vaughan]{ben2010theory}
Shai Ben-David, John Blitzer, Koby Crammer, Alex Kulesza, Fernando Pereira, and
  Jennifer~Wortman Vaughan.
\newblock A theory of learning from different domains.
\newblock \emph{Machine learning}, 79\penalty0 (1-2):\penalty0 151--175, 2010.

\bibitem[Stephens(1974)]{stephens1974edf}
Michael~A Stephens.
\newblock Edf statistics for goodness of fit and some comparisons.
\newblock \emph{Journal of the American statistical Association}, 69\penalty0
  (347):\penalty0 730--737, 1974.

\bibitem[Newby et~al.(2012)Newby, Jesse, Babb, Christenson, De~Fer, Diamond,
  Fesmire, Geraci, Gersh, Larsen, et~al.]{newby2012accf}
L~Kristin Newby, Robert~L Jesse, Joseph~D Babb, Robert~H Christenson, Thomas~M
  De~Fer, George~A Diamond, Francis~M Fesmire, Stephen~A Geraci, Bernard~J
  Gersh, Greg~C Larsen, et~al.
\newblock Accf 2012 expert consensus document on practical clinical
  considerations in the interpretation of troponin elevations: a report of the
  american college of cardiology foundation task force on clinical expert
  consensus documents.
\newblock \emph{Journal of the American College of Cardiology}, 60\penalty0
  (23):\penalty0 2427--2463, 2012.

\bibitem[Chen and Guestrin(2016)]{Chen:2016:XST:2939672.2939785}
Tianqi Chen and Carlos Guestrin.
\newblock {XGBoost}: A scalable tree boosting system.
\newblock In \emph{Proceedings of the 22nd ACM SIGKDD International Conference
  on Knowledge Discovery and Data Mining}, KDD '16, pages 785--794, New York,
  NY, USA, 2016. ACM.
\newblock ISBN 978-1-4503-4232-2.
\newblock \doi{10.1145/2939672.2939785}.
\newblock URL \url{http://doi.acm.org/10.1145/2939672.2939785}.

\bibitem[Rajpurkar et~al.(2017)Rajpurkar, Hannun, Haghpanahi, Bourn, and
  Ng]{rajpurkar2017cardiologist}
Pranav Rajpurkar, Awni~Y Hannun, Masoumeh Haghpanahi, Codie Bourn, and Andrew~Y
  Ng.
\newblock Cardiologist-level arrhythmia detection with convolutional neural
  networks.
\newblock \emph{arXiv preprint arXiv:1707.01836}, 2017.

\end{thebibliography}
}

\clearpage
\appendix

\clearpage
\section{Additional Figures}
\begin{figure}[h!]
    \centering
    \begin{subfigure}[b]{.49\linewidth}
        \centering
        \includegraphics[width=\textwidth]{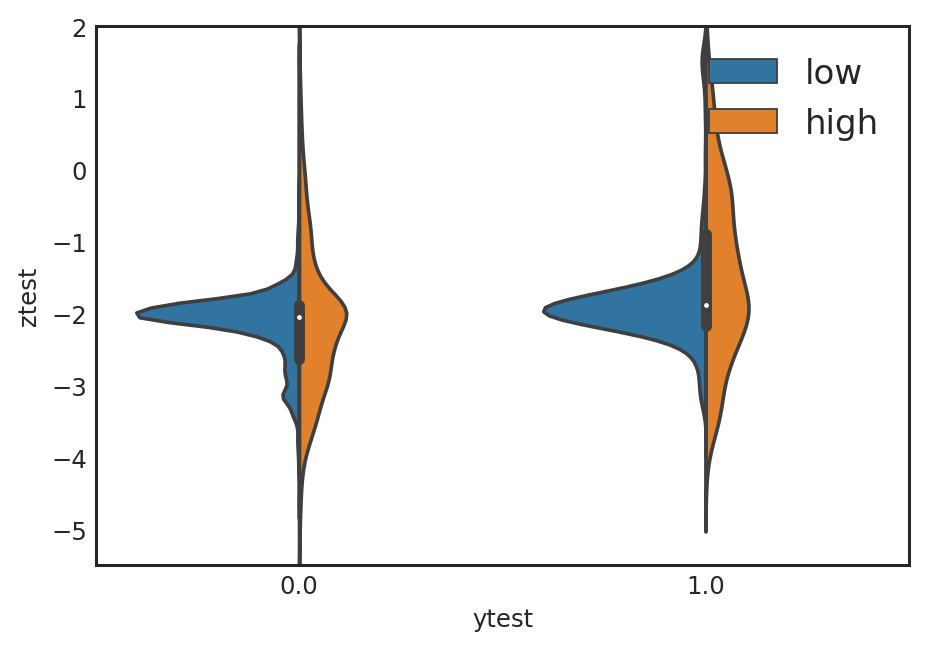}
        \caption{\texttt{ehr}, trained on low-use ($Q$)}
    \end{subfigure}%
    \begin{subfigure}[b]{.49\linewidth}
        \centering
        \includegraphics[width=\textwidth]{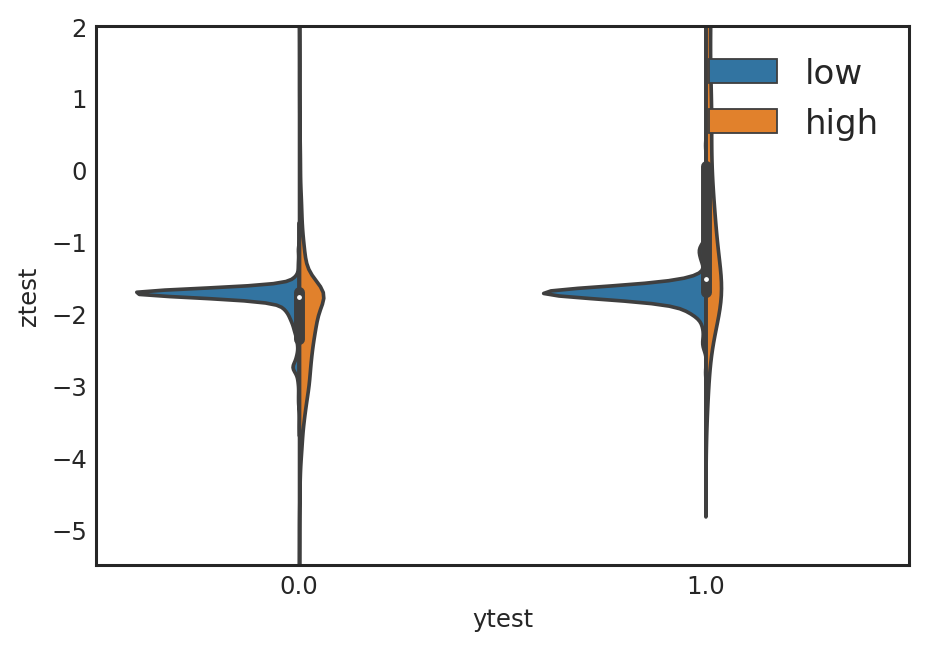}
        \caption{\texttt{ehr}, trained on high-use ($P$) }
    \end{subfigure}
    
    \begin{subfigure}[b]{.49\linewidth}
        \centering
        \includegraphics[width=\textwidth]{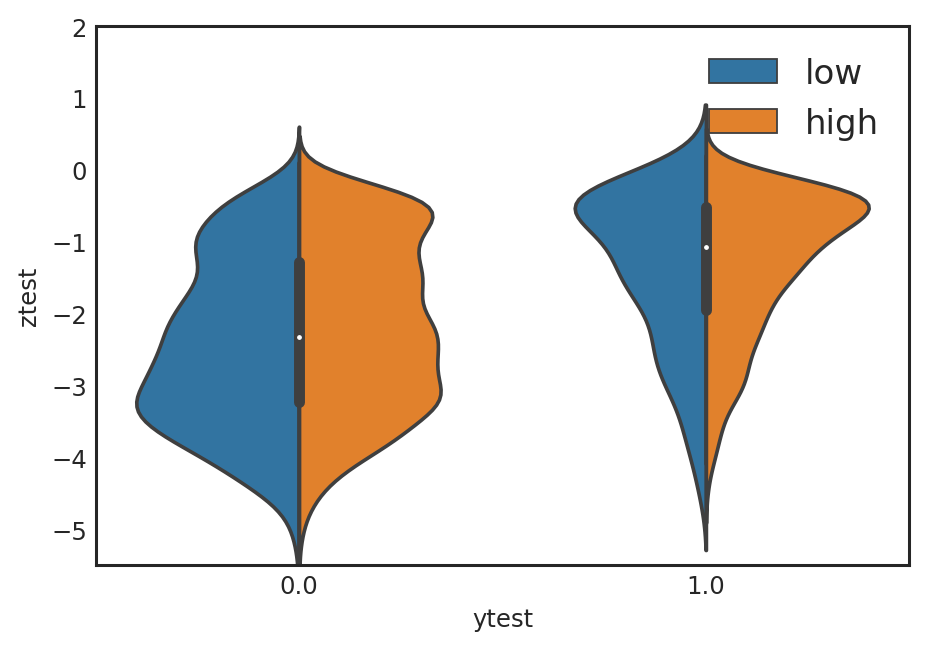}
        \caption{\texttt{ekg}, trained on low-use ($Q$) }
    \end{subfigure}%
    \begin{subfigure}[b]{.49\linewidth}
        \centering
        \includegraphics[width=\textwidth]{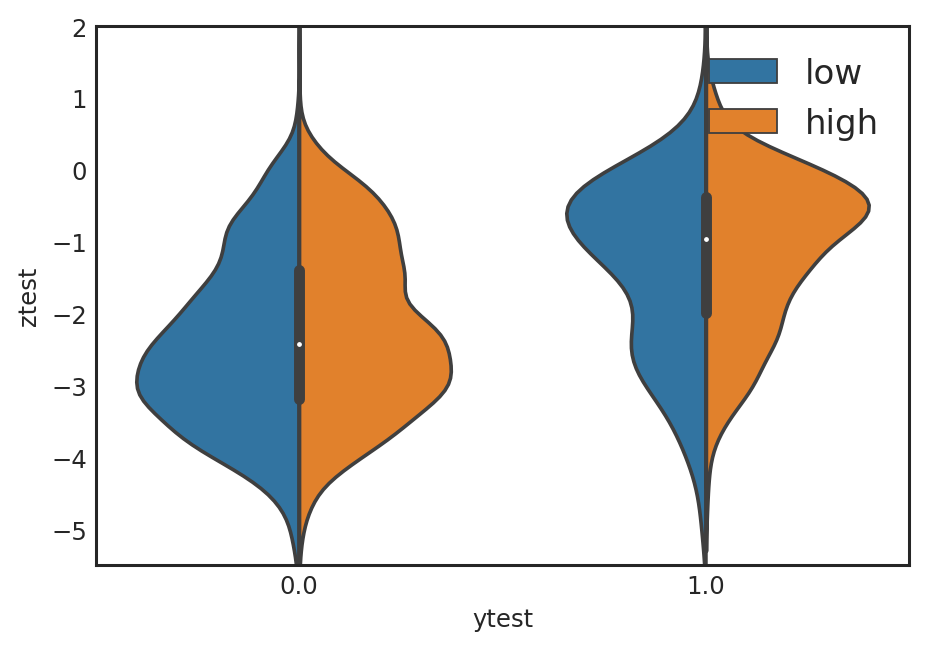}
        \caption{\texttt{ekg}, trained on high-use ($Q$) }
    \end{subfigure}%
    \caption{Covariate stability.  Above we visualize the conditional distribution of $\hat y$ given the true label $y=0$ or $y=1$ (denoted along the horizontal axis).  
    Within the $y=0$ or $y=1$ subset, the blue distributions denote the predictive $\hat y$ distribution among the $Q=$ low use patients, and the orange denotes the predictive $\hat y$ distribution among the $P=$ high use patients. 
    Intuitively, a stable predictor would have a similar distribution of $\hat y$ across  $P$ and $Q$ (but within the group true $y=1$ or $y=0$). 
    For example (a) shows the conditional distribution of $\hat y | y=0$ for the $Q$ (blue) and $P$ (orange) populations, for $\hat y$ trained on \texttt{ehr} features from population $Q$. (b) depicts the same conditional distributions for \texttt{ehr} features trained on $P$.  (c) and (d) depict the same breakdown for models trained on \texttt{ekg} features.  
    The covariates are ``stable'' when the conditional distributions match --- we see the \texttt{ekg}-based covariates are much more stable than \texttt{ehr}-based covariates.  This similarity is summarized by the covariate similarity statistic Figure~\ref{fig:covariate-stability}.  }
    \label{fig:covariate-distributions}
\end{figure}

\begin{figure}[h!]
    \centering
    \begin{subfigure}[b]{.49\linewidth}
        \centering
        \includegraphics[width=\textwidth]{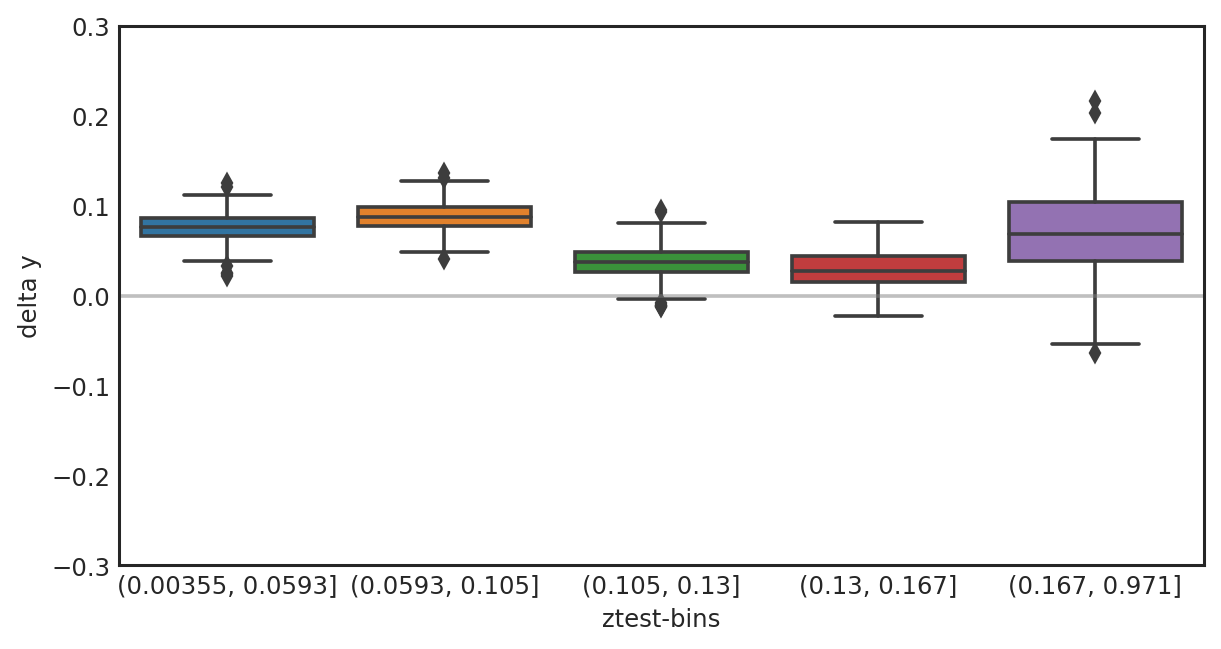}
        \caption{\texttt{ekg} trained on $Q$}
    \end{subfigure}%
    \begin{subfigure}[b]{.49\linewidth}
        \centering
        \includegraphics[width=\textwidth]{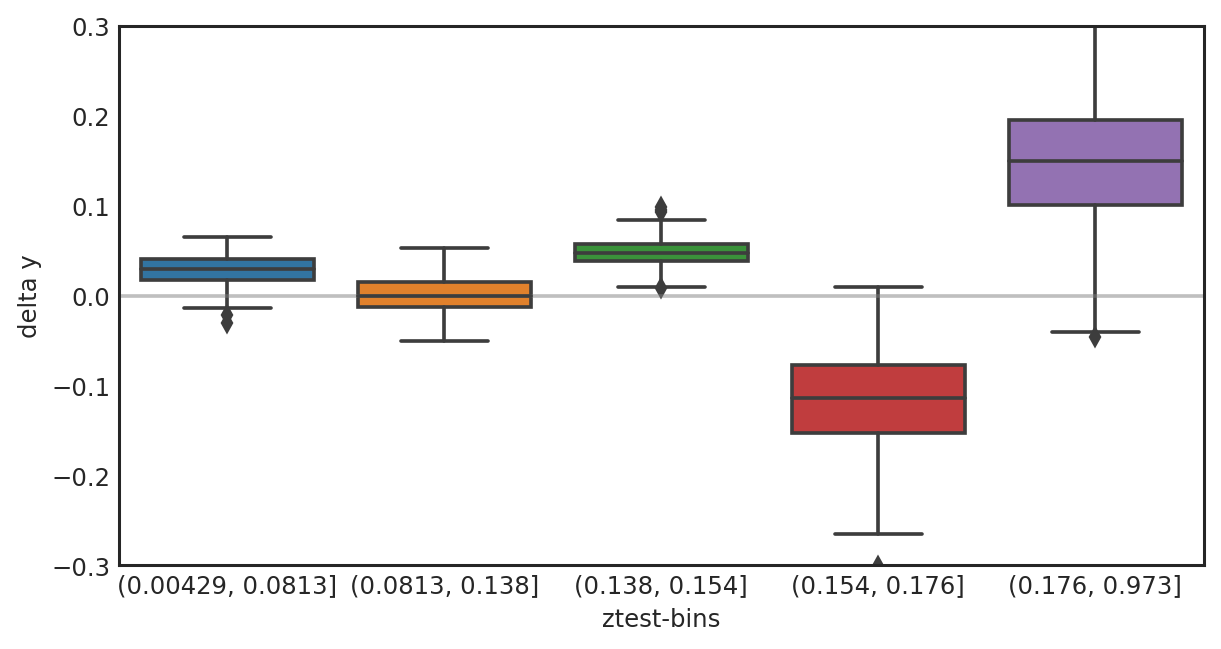}
        \caption{\texttt{ekg} trained on $P$}
    \end{subfigure}
    
    \begin{subfigure}[b]{.49\linewidth}
        \centering
        \includegraphics[width=\textwidth]{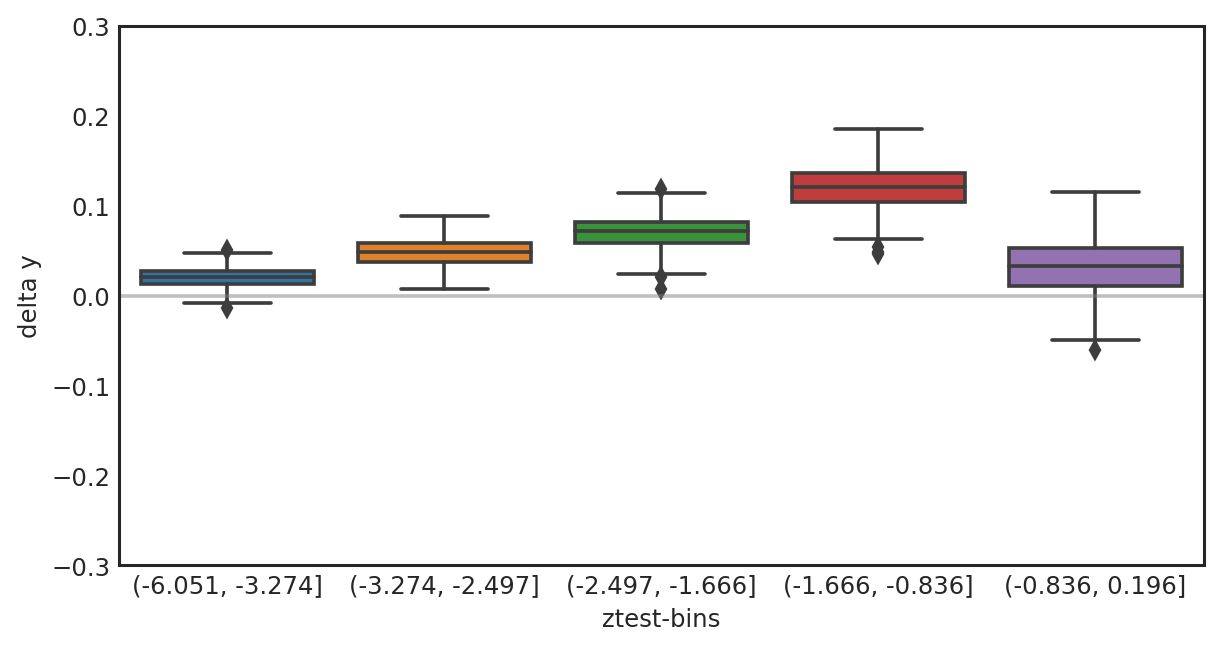}
        \caption{}
    \end{subfigure}%
    \begin{subfigure}[b]{.49\linewidth}
        \centering
        \includegraphics[width=\textwidth]{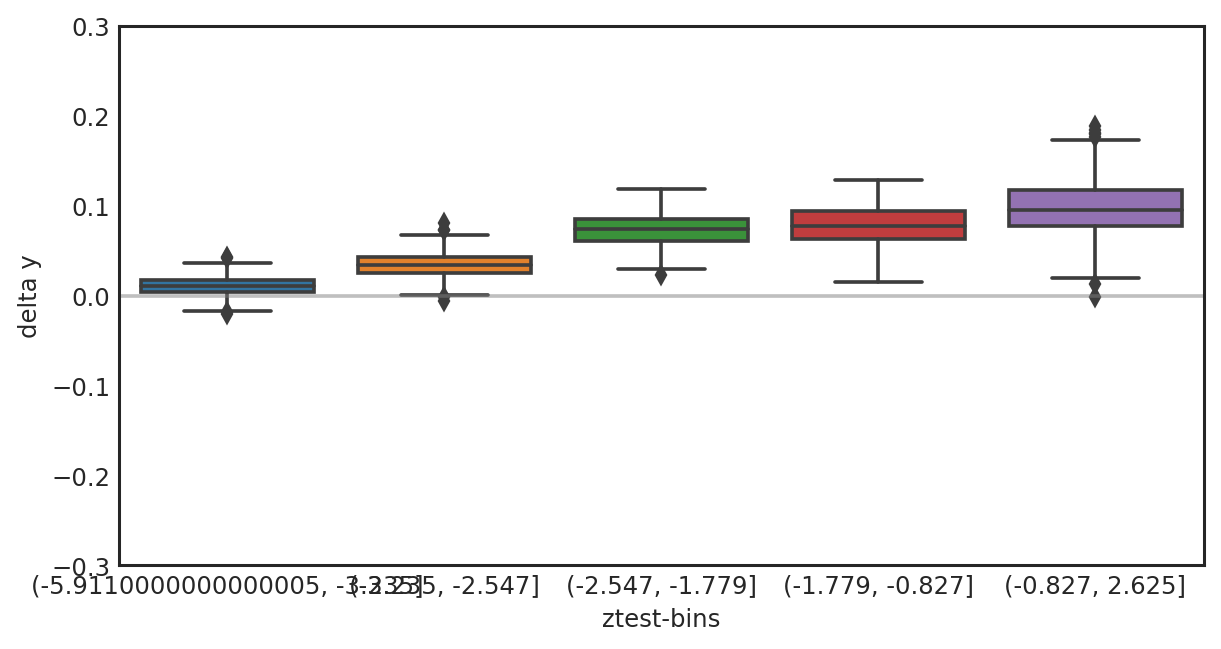}
        \caption{}
    \end{subfigure}%
    \caption{Predictive stability.  Here we depict estimates of the predictive stability curve described in Equation~\ref{eq:predictive-stability}.  Each bin along the horizontal axis reflects a quintile of test data, stratified by $\hat y$. 
    Here we see a more complete picture --- the \texttt{ehr}-based predictors tend to be quite predictively stable.  The \texttt{ekg}-based predictors appear to be slightly less stable, which may indicate slight differences in the physiology of low-use $Q$ and high-use $P$ patients (for instance, high-use patients are on average a bit older and male in this sample). 
    }
    \label{fig:predictive-stability-full}
\end{figure}

\end{document}